\documentclass[conference]{IEEEtran}
\IEEEoverridecommandlockouts
\usepackage{cite}
\usepackage{amsmath,amssymb,amsfonts}
\usepackage{algorithmic}
\usepackage{graphicx}
\usepackage{textcomp}
\usepackage{xcolor}
\usepackage{enumitem}
\usepackage{booktabs} 
\usepackage{makecell}
\usepackage{multirow}
\usepackage{siunitx} 

\def\BibTeX{{\rm B\kern-.05em{\sc i\kern-.025em b}\kern-.08em
    T\kern-.1667em\lower.7ex\hbox{E}\kern-.125emX}}
\begin{document}

\title{Reproducible Evaluation of Data Augmentation and Loss Functions for Brain Tumor Segmentation
\\
}

\author{\IEEEauthorblockN{ Saumya B}
\IEEEauthorblockA{\textit{Project Associate} \\
\textit{DESE, Indian Institute of Science}\\
Bengaluru, India \\
saumya.b@fsid-iisc.in}

}

\maketitle

\begin{abstract}
Brain tumor segmentation is crucial for diagnosis and treatment planning, yet challenges such as class imbalance and limited model generalization continue to hinder progress. This work presents a reproducible evaluation of U-Net segmentation performance on brain tumor MRI using focal loss and basic data augmentation strategies. Experiments were conducted on a publicly available MRI dataset, focusing on focal loss parameter tuning and assessing the impact of three data augmentation techniques: horizontal flip, rotation, and scaling. The U-Net with focal loss achieved a precision of 90\%, comparable to state-of-the-art results. By making all code and results publicly available, this study establishes a transparent, reproducible baseline to guide future research on augmentation strategies and loss function design in brain tumor segmentation.
\end{abstract}


\section{Introduction}
Brain tumors are among the most challenging medical conditions to diagnose and treat, often requiring precise identification of tumor boundaries for effective treatment planning. Magnetic Resonance Imaging (MRI) is a widely used imaging modality for detecting brain tumors, providing detailed anatomical information essential for accurate diagnosis. However, manual delineation of tumor regions by radiologists is time-consuming, prone to inter-observer variability, and difficult to scale in clinical settings. These challenges highlight the need for automated methods that can accurately segment brain tumors and provide interpretable predictions.

Deep learning-based approaches, particularly convolutional neural networks (CNNs), have shown significant promise in medical image segmentation tasks due to their ability to learn hierarchical spatial features. Among these, the U-Net architecture is widely regarded as the gold standard for biomedical image segmentation, owing to its encoder-decoder structure with skip connections that effectively capture fine-grained spatial details \cite{b1}.While U-Net models have achieved strong segmentation performance, they continue to face challenges with class imbalance, a factor that can significantly affect model accuracy in medical imaging.

This study presents a reproducible evaluation of U-Net performance on brain tumor MRI with focal loss and basic data augmentation strategies. Focal loss is commonly applied to address class imbalance by increasing the weight of hard-to-classify regions, such as tumor boundaries. We systematically evaluated the effect of focal loss parameters alongside three augmentation techniques — horizontal flip, rotation, and scaling — to assess their impact on segmentation accuracy and generalization.

The major contributions of this work are given below:
\begin{itemize}
 \item A baseline implementation of U-Net with focal loss for brain tumor MRI segmentation 
\item Analysis of the impact of varying focal loss parameters on model performance
 \item Evaluation of the effect of three different data augmentation techniques on model performance
 \end{itemize}
Though this study does not introduce new architectures, it provides a transparent, reproducible evaluation of focal loss and data augmentation strategies for brain tumor segmentation. By making the results and code publicly available, this work serves as a reproducible baseline and reference point for future research in model training, augmentation strategies, and loss function design.

\section{Literature Survey}
Accurate quantification of brain tumors is critical for diagnosis, treatment planning, and monitoring of therapeutic response. Among imaging techniques, MRI remains the preferred modality because of its superior soft tissue contrast, enabling more precise tumor delineation than CT or ultrasound. Traditional segmentation methods, such as thresholding \cite{b2, b3}, boundary detection via active contour models \cite{b4}, and various forms of region growing \cite{b5, b6, b7}, depend on handcrafted intensity heuristics \cite{b7} or feature engineering \cite{b5}. For example, Sujan et al. (2016) \cite{b3} combined thresholding with morphological operations, reporting approximately 85\% accuracy for tumor detection on BRATS MRI scans, while Meier et al. (2016) \cite{b8} demonstrated clinical validation of a fully automated volumetry system in longitudinal studies. Nevertheless, such rule-based or classical approaches are frequently constrained by noise sensitivity and poor generalization across datasets, thus motivating the shift toward deep learning-based segmentation techniques.

Convolutional neural networks (CNNs) have transformed medical image analysis by automatically learning hierarchical features, outperforming traditional handcrafted approaches in tumor detection and segmentation \cite{b9}. Among these, U-Net (Ronneberger et al., 2015) \cite{b1} became the standard for biomedical segmentation due to its encoder–decoder structure with skip connections, enabling precise localization from limited training data. However, brain tumor segmentation remains challenging because of severe class imbalance, where tumor pixels are sparse compared to the background. To address this, several loss functions have been explored: binary cross-entropy (BCE) provides pixel-level supervision but struggles with imbalance, while Dice loss emphasizes overlap but can be unstable for small structures. More recently, focal loss \cite{b10} has been proposed to focus training on hard-to-classify pixels, and studies comparing focal loss with BCE demonstrate its advantage in medical segmentation tasks \cite{b11}.

Data scarcity and variability across scanners often limit the generalization of brain tumor segmentation models. Data augmentation is therefore widely applied to synthetically enlarge training sets and improve robustness. Common techniques include geometric transformations such as flipping, rotation, and scaling, as well as elastic deformations and intensity perturbations \cite{b12}. Prior studies have shown that even simple augmentations can significantly boost Dice scores and reduce overfitting in medical segmentation tasks. Since data augmentation is already known to improve brain tumor segmentation, this study focuses on systematically comparing different augmentation strategies in this context.

\section{Methodology}
To achieve the objectives mentioned in section 1, the study was conducted in two phases, using the same U-Net model:

\begin{enumerate}[label=\roman*.] 

\item \textit{Focal Loss Parameter Tuning:}

In the initial set of experiments, the goal was to identify the optimal parameters for focal loss that yielded the best segmentation performance. These experiments were conducted using the original dataset without applying any data augmentation techniques.
\item \textit{Data Augmentation Analysis:}

After determining the best-performing focal loss parameters, they were fixed for the subsequent experiments. Three different data augmentation techniques were then applied individually to the dataset, and the model’s performance was evaluated for each augmentation technique.
\end{enumerate}
This systematic approach ensured a comprehensive analysis of both focal loss parameter tuning and the impact of data augmentation on the model’s segmentation performance.

\subsection{Dataset}
For this study, a brain tumor dataset containing 3064 T1- weighted contrast-enhanced MRI images was used. The data was collected from Nanfang Hospital and Tianjing Medical University, China, from 2005 to 2010, by Jun Cheng, who originally used the dataset in his study \cite{b13} and \cite{b14}, and had uploaded the entire dataset with its metadata to Figshare. The dataset consists of T1-weighted contrast-enhanced MRI scans from 233 patients, and has three kinds of brain tumors - 708 cases of Meningiomas, 1426 gliomas and 930 cases of pituitary tumors. The source states that the tumor borders were manually delineated by three experienced radiologists. A copy of this dataset was taken from Kaggle \cite{b15}, where the scans and corresponding binary masks were uploaded as 256x256 pixel images. A summary of the details of the dataset are given in Table 1.

\begin{table}[h]
    \centering
    \caption{Dataset details}
    \label{tab:my_simple_table}
    \begin{tabular}{l c r}
        \hline
        \textbf{Tumor type} & \textbf{Dimensions} & \textbf{No. of images} \\
        \hline
         Meningioma & 256x256 & 708 \\
        \hline
        Glioma & 256x256 & 1426 \\
        \hline
        Pituitary & 256x256 & 930 \\
        \hline
    \end{tabular}
\end{table}

\subsection{Pre-processing}
\begin{figure}[htbp]
\centerline{\includegraphics[scale = 0.5]{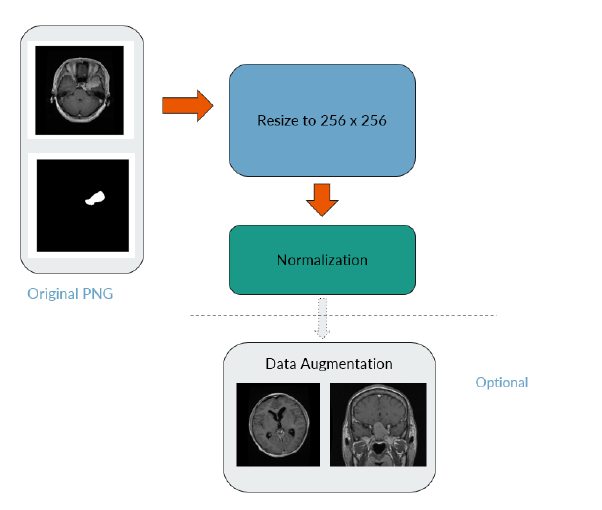}}
\caption{Pre-processing workflow}
\label{fig}
\end{figure}
The preprocessing steps began by loading both the images and masks in grayscale format, hence reducing the complexity by using fewer channels compared to color images. The images and masks were then resized to a standardized size (256x256) to ensure uniformity across the dataset. Following this, the pixel values were normalized by dividing each pixel by 255, which not only facilitates faster convergence \cite{b16} during training but also makes the computations more efficient and less memory-intensive. Since the model is designed to work with binary masks, an important step was performed to verify that all mask pixel values were either 0 or 1. Any pixel with a value between 0 and 1 (non-binary) was converted to 0 if it was greater than 0 but less than 1. Finally, the dataset, which contained 3064 samples, was randomly split into training, validation, and test sets, with 1838 samples assigned to the training set, and 613 samples each assigned to the validation and test sets.

\subsection{Model Architecture}

The model used in the study is the U-Net architecture, originally designed for biomedical image segmentation tasks. The U-Net architecture was selected due to its effectiveness in biomedical image segmentation, particularly in tasks with limited labeled data and the need for precise spatial localization, which is critical for tumor segmentation. It employs an encoder-decoder structure with skip connections to preserve spatial information. The encoder extracts hierarchical features through downsampling, while the decoder reconstructs the segmentation mask by upsampling the feature maps.

The encoder consists of four downsampling blocks, each with two convolutional layers (with padding = “same”, kernel size of 3x3) to balance the need for fine-grained segmentation accuracy and model generalization. This is followed by a max pooling layer with a pool size of 2x2. The convolutional layers use ReLU activation and He normal initialization. ReLU was chosen because it introduces non-linearity, avoids vanishing gradient issues, and promotes faster convergence, while He initialization complements it by maintaining variance throughout the network, ensuring stable and efficient training. After the pooling layers, a dropout layer with a rate of 0.3 is applied to regularize the network and prevent overfitting. The number of filters doubles in each block, starting from 64 in the first block and increasing to 1024 in the bottleneck layer.

The decoder mirrors the encoder with four upsampling blocks. Each block begins with a transposed convolution layer to upsample the feature maps, followed by skip connections that concatenate feature maps from the corresponding encoder block, preserving spatial details lost during down-sampling. These are followed by two convolutional layers with kernel size 3x3 and with ReLU activation and He initialization. A dropout layer with the same rate of 0.3 is applied after the concatenation step to maintain regularization and support uncertainty estimation. At the bottleneck, the deepest part of the U-Net, two convolutional layers with 1024 filters capture high-level feature representations essential for accurate segmentation. The output layer is a 1×1 convolution with a sigmoid activation function, producing a binary segmentation map to identify tumor regions.

\subsection{Choosing focal loss as loss function}
Focal loss was selected as the loss function for this model due to its effectiveness in addressing class imbalance, which is a common issue in segmentation tasks. In this case, the segmentation masks contain significantly fewer foreground pixels (tumor regions) compared to the background pixels. Standard loss functions, such as cross-entropy loss, may struggle in such scenarios because the loss tends to be dominated by the majority class (background). 

Focal Loss mitigates this issue by dynamically scaling the contribution of each pixel’s loss based on its classification difficulty. Hard-to-classify examples (e.g., noisy textures, small tumor regions, or partial objects) are given higher weights, while easy-to-classify examples (e.g., background pixels) are down-weighted, allowing the model to focus on learning challenging patterns. Focal Loss is an extension of Cross-Entropy Loss, formulated as in Eq. \ref{eq:focal_loss}.
\begin{equation}
    \label{eq:focal_loss}
    FL(p_t) = -\alpha (1 - p_t)^\gamma \log(p_t)
\end{equation}
where $p_t$ is the model’s predicted probability for the true class, $\alpha$ is a weighting factor for balancing the importance of different classes, (1 - $p_t$) is the modulating factor, which gives more weight to misclassified or hard examples, and $\gamma$ is the focusing parameter that adjusts how much the loss focuses on hard examples. The parameters $\alpha$ and $\gamma$ play crucial roles in the behavior of Focal Loss. $\alpha$ is the weighting factor which controls the balance between the foreground and background classes. For instance, setting a higher $\alpha$ for the tumor (foreground) class ensures that the model assigns greater importance to learning from tumor pixels, which are typically underrepresented. $\gamma$ is the focusing parameter and determines the degree of focus on hard examples. When  $\gamma$ = 0, Focal Loss is equivalent to standard Cross-Entropy Loss. Increasing $\gamma$ (e.g., 2.0) amplifies the contribution of harder examples while reducing the weight of well-classified ones. Higher values of  $\gamma$ are particularly useful in datasets with severe class imbalance.

\subsection{Hyperparamenters}

\begin{table}[h!]
    \centering
    \caption{Model Hyperparameters and Training Configuration}
    \label{tab:model_config}
    \begin{tabular}{p{3cm} l l}
        \toprule
        \textbf{Category} & \textbf{Hyperparameter} & \textbf{Value} \\
        \midrule
        \multirow{8}{*}{\textbf{Model Architecture}} & Input Shape & (256, 256, 1) \\
        & Kernel Size (Initial) & (3, 3) \\
        & Kernel Size (Subsequent) & (3, 3) \\
        & Activation function & ReLU \\
        & Kernel Initializer & He Normal \\
        & Dropout Rate & 0.3 \\
        & Final Activation & Sigmoid \\
        \midrule
        \multirow{3}{*}{\textbf{Loss and Metrics}} & Loss function & Focal Loss \\
        & Optimizer & Adam \\
        & Training metrics & Accuracy \\
        \midrule
        \multirow{3}{*}{\textbf{Training Parameters}} & Batch Size & 8 \\
        & Learning Rate & $1 \times 10^{-4}$ \\
        & No. of epochs & 200 \\
        \bottomrule
    \end{tabular}
\end{table}

The Adam optimizer (Adaptive Momentum Estimator) was chosen for training the model due to its effectiveness in biomedical image segmentation tasks \cite{b17} and its ability to balance computational efficiency, minimal memory requirements, and ease of implementation. Unlike traditional stochastic gradient descent, Adam adapts the learning rate dynamically for each parameter based on past gradients, making it well-suited for complex models like U-Net. The learning rate is a critical hyperparameter that directly affects training stability and speed. A learning rate that is too low can result in slow convergence, while one that is too high may cause the model to miss the optimal solution. For this model, an initial learning rate of $1 \times 10^{-4}$
 was selected, which provided a good balance between convergence speed and stability. The rest of the hyperparameters chosen are given in detail in the Table 2.

\subsection{Metrics}
The evaluation was focused on assessing segmentation quality. Segmentation quality was assessed using standard metrics such as Dice Coefficient, Intersection Over Union (IoU), precision and recall. These metrics provide a quantitative measure of the overlap and accuracy of the predicted segmentation masks compared to the ground truth.

\subsection{Training}
All training was performed using Google Colab’s free TPU, the TPUv2-8. The Tensorflow library was used to enable the model to utilise the TPU during training. The model was trained for 200 epochs using Adam optimizer with a learning rate of $1 \times 10^{-4}$. A fixed training schedule of 200 epochs was used across all experiments to ensure consistency and to analyze when overfitting tendencies emerged. While early stopping or best-model checkpointing are commonly applied, in this study the extended training allowed us to observe the effect of different loss parameters and augmentations on convergence and stability. Validation was done using the standard training-validation-test split of 60\%-20\%-20\%. During training, accuracy was used as the primary evaluation metric, which provides a general measure of how well the model’s predictions align with the ground truth. However, for a more comprehensive evaluation of segmentation performance, additional metrics were computed on the test dataset, including the Dice coefficient, Intersection over Union (IoU), precision, and recall. These metrics are particularly relevant for assessing the quality of segmentation masks, especially in imbalanced datasets where pixel-level accuracy may not fully reflect model performance.

\subsection{Experiments conducted}
\textit{Phase 1: Focal Loss Parameter Tuning}

In this phase of the experiments, the $\alpha$ and $\gamma$ parameters for focal loss were varied to evaluate their impact on segmentation performance. These experiments were conducted using the original dataset, without applying any data augmentation techniques. Two sets of parameter combinations were tested:
\begin{list}{}{%
    \setlength{\itemindent}{1em} 
    \setlength{\itemsep}{0pt}    
}
\item[\textbullet] $\alpha=0.25, \gamma=2.0$
\item[\textbullet] $\alpha=2.0, \gamma=0.75$
\end{list}
For the first set of parameters ($\alpha$ = 0.25, $\gamma$ = 2.0), the lower $\alpha$ places more weight on the background class, while the higher $\gamma$ emphasizes hard-to-classify pixels, such as tumor boundaries or ambiguous regions. In contrast, the second set of parameters ($\alpha$ = 2.0,$\gamma$ = 0.75) increases the importance of the minority class (tumor) with a higher $\alpha$, while the smaller $\gamma$ of 0.75 reduces the focus on difficult examples and instead prioritizes the overall distribution, giving weightage to easier- to-classify regions too.

\textit{
   Phase 2: Data Augmentation Evaluation}

In this phase, three different data augmentation techniques were evaluated on the proposed model, while keeping the focal loss parameters constant at $\alpha$ = 0.25 and $\gamma$ = 2.0. The techniques evaluated are: Horizontal Flip, Rotation and Scaling.
The details of these augmentation techniques are provided in Table 3. It is important to note that, since this is a segmentation task, the augmentation transformations were applied to both the image and its corresponding mask to ensure consistency between the input and target.
\begin{table}[h]
    \centering
    \caption{Data Augmentation Experiment details}
    \label{tab:data_augmentation}
    \begin{tabular}{l c l}
        \toprule
        \textbf{Technique} & \textbf{\% of training dataset} & \textbf{Parameters} \\
        \midrule
        Horizontal Flip & 50\% & none \\
        Rotation & 50\% & Angle: \SI{+-15}{\degree} \\
        Random Scaling & 50\% & Range: 0.8 - 1.2 \\
        \bottomrule
    \end{tabular}
\end{table}
\section{Results and discussion}
This section presents the results of the experiments conducted to evaluate the segmentation performance of the model, the impact of focal loss parameter tuning and the effect of data augmentation techniques. The model was evaluated using multiple segmentation metrics, including accuracy, loss, Dice coefficient, Intersection over Union (IoU), precision, and recall. Experiments were conducted using both the original dataset and augmented datasets to assess the impact of data augmentation techniques on performance. The model showed competitive results across all metrics, with Horizontal Flip emerging as the most effective data augmentation technique, achieving the highest Dice coefficient and IoU scores. Rotation also contributed positively, whereas Scaling had minimal impact on model performance.

\subsection{Impact of focal loss parameters}
The results obtained in the first set of experiments of tuning the focal loss are presented in Table 4. In the first case, the configuration prioritized hard-to-classify examples, particularly tumor boundaries, resulting in better performance. In the second case, the configuration emphasized the minor class (tumor regions) more but focused less on boundary details, leading to weaker performance in challenging regions.

The experiments demonstrated that the choice of focal loss parameters play a significant role in balancing foreground and background segmentation accuracy.

\begin{table*}[htbp]
    \centering
    \caption{Results for Loss Parameter Experiments}
    \label{tab:loss_results}
    \begin{tabular}{c c S[table-format=1.4] S[table-format=1.4] S[table-format=1.4] S[table-format=1.4] S[table-format=1.4] S[table-format=1.4]}
        \toprule
        \textbf{Loss Function} & \textbf{Parameters} & {\textbf{Accuracy}} & {\textbf{Loss}} & {\textbf{Precision}} & {\textbf{Recall}} & {\textbf{IoU}} & {\textbf{Dice Co-efficient}} \\
        \midrule
        \multirow{2}{*}{Focal Loss} & \makecell{$\alpha=0.25$ \\ $\gamma=2.0$} & 0.9941 & 0.0082 & 0.9014 & 0.7681 & 0.7082 & 0.7867 \\
        \cmidrule{2-8} 
        \multirow{2}{*}{Focal Loss} & \makecell{$\alpha=2.0$ \\ $\gamma=0.75$} & 0.9939 & 0.0154 & 0.8778 & 0.7789 & 0.7004 & 0.7839 \\
        \bottomrule
    \end{tabular}
\end{table*}

\begin{table*}[htbp]
    \centering
    \caption{Results for Data Augmentation Experiments}
    \label{tab:augmentation_results}
    
    \begin{tabular}{l *6{S[table-format=1.4]}} 
        \toprule
        \textbf{Augmentation type} & {\textbf{Accuracy}} & {\textbf{Loss}} & {\textbf{Precision}} & {\textbf{Recall}} & {\textbf{IoU}} & {\textbf{Dice Co-efficient}} \\
        \midrule
        None & 0.9941 & 0.0082 & 0.9014 & 0.7681 & 0.7082 & 0.7867 \\
        Horizontal Flip & 0.9942 & 0.0053 & 0.9001 & 0.7779 & 0.7152 & 0.8041 \\
        Rotation & 0.9940 & 0.0029 & 0.8774 & 0.7892 & 0.7090 & 0.7955 \\
        Random Scaling & 0.9934 & 0.0064 & 0.9097 & 0.7106 & 0.6643 & 0.7486 \\
        \bottomrule
    \end{tabular}
\end{table*}

\subsection{Impact of Data Augmentation}
Three data augmentation techniques - Horizontal flip, Rotation and Scaling - were experimented with to evaluate their effect on model generalization and robustness. The results of the experiments are summarized in the Table 5. Horizontal flip consistently improved performance across all metrics, making it the most effective augmentation technique. Rotation improved the Dice coefficient and IoU, indicating its effectiveness. Scaling, however, showed negligible or no improvement in segmentation performance, suggesting that size variations are less significant for this dataset. Visualization of ground truth and predictions further supported these findings (refer figure 2), with horizontal flip and rotation demonstrating clearer and more accurate segmentation results compared to scaling.
\setlength{\tabcolsep}{2pt} 

\begin{figure}[htbp]
\centerline{\includegraphics[scale=0.25]{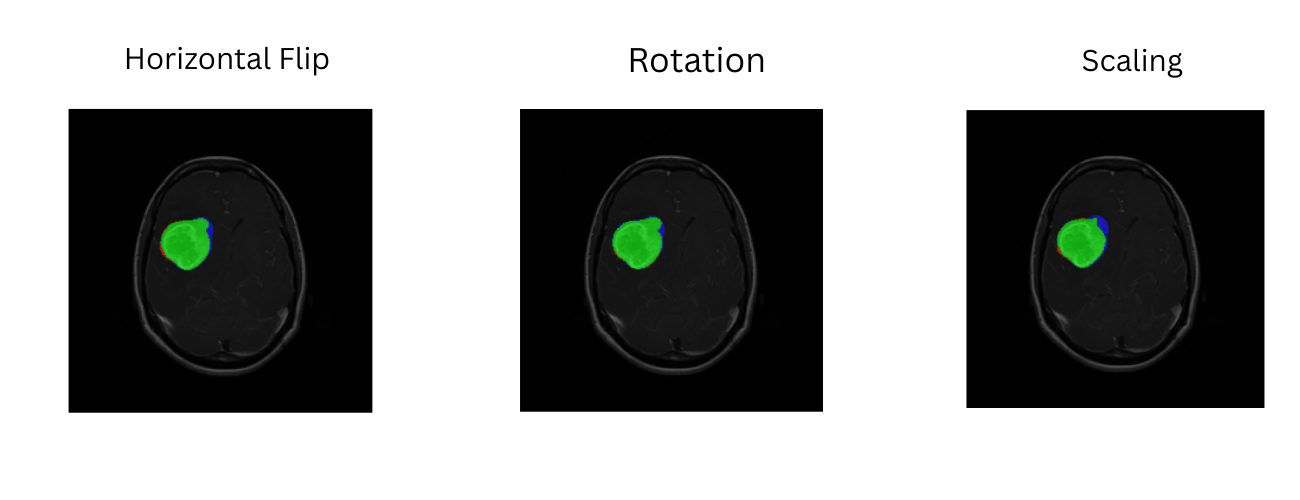}}
\caption{Ground Truth v/s Model Prediction: Green - Prediction, Blue - GT}
\label{fig}
\end{figure}
\subsection{Inference drawn from Training graphs}
\begin{figure*}[htbp]
\centerline{\includegraphics[scale = 0.33]{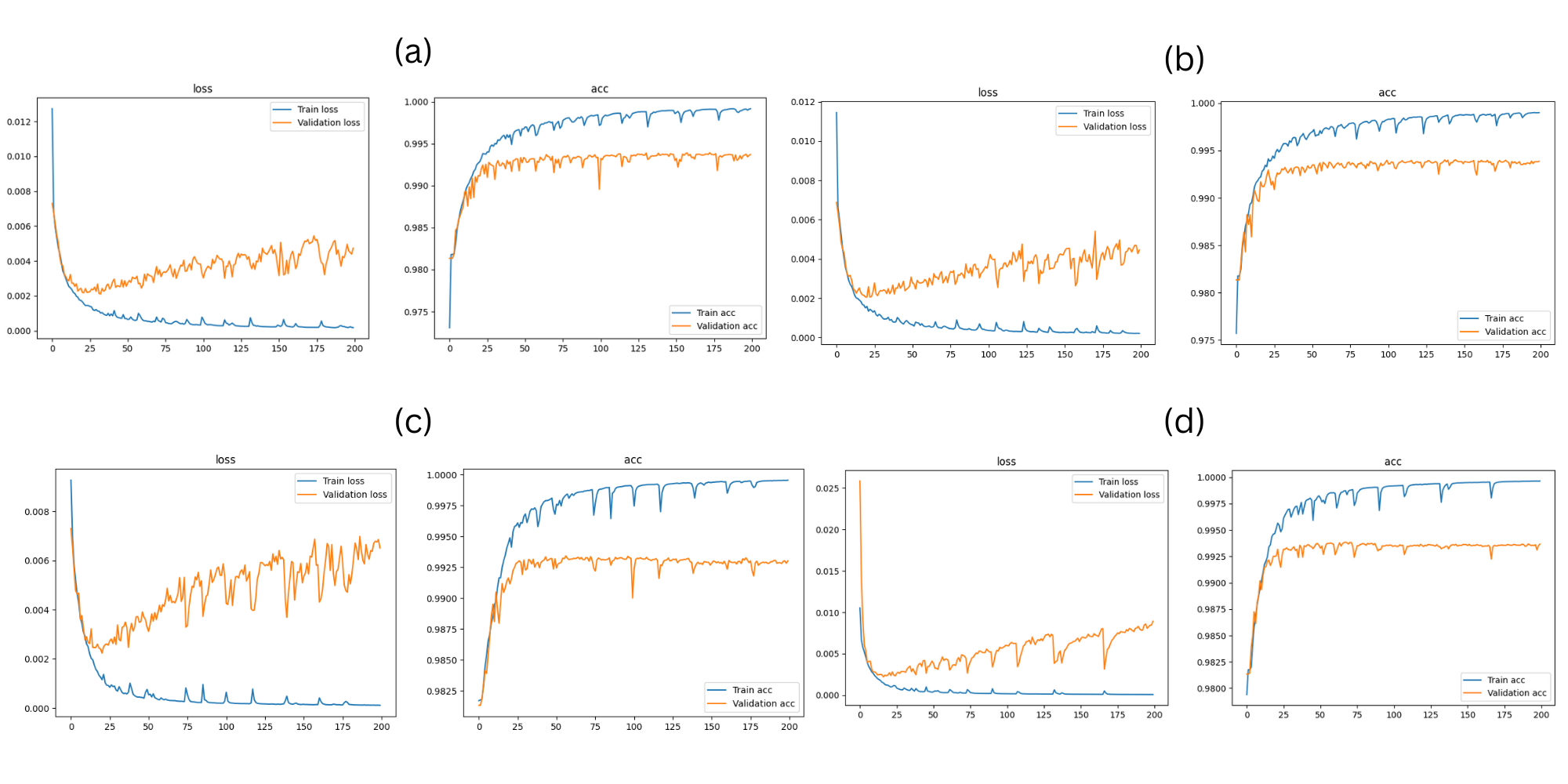}}
\caption{Training graphs - (a) Horizontal flip (b) Rotation (c) Scaling (d) No aug; blue - train, orange - validation}
\label{fig}
\end{figure*}
Figure 3 shows the training graphs for the different augmentation techniques compared with the baseline model. Horizontal flip and rotation augmentations produced more stable validation curves with smaller gaps between training and validation performance, suggesting better robustness to unseen data. In contrast, scaling augmentation resulted in higher fluctuations in validation loss, indicating reduced stability and weaker generalization. The no-augmentation setting showed smoother curves but a clear gap between training and validation, pointing to mild overfitting. Overall, the results suggest that simple geometric augmentations such as flip and rotation contribute more effectively to generalization compared to scaling or no augmentation. Extending training to 200 epochs further highlighted these differences, showing that horizontal flip consistently accelerated convergence with stable behavior, rotation offered moderate but steady gains, while scaling produced noisy training patterns and minimal benefit. 
\subsection{Comparison with state-of-the-art}
The model achieved comparable performance with state-of-the-art methods for brain tumor segmentation. A brief comparison with state-of-the-art models is shown in Table 7.
\begin{table}[h!] 
    \centering
    \caption{Comparison with State-of-the-Art (SOTA)}
    \label{tab:sota_comparison}
    
    \begin{tabular}{l *4{S[table-format=1.4]}} 
        \toprule
        \textbf{Model} & {\textbf{Precision}} & {\textbf{Recall}} & {\textbf{IoU}} & {\textbf{Dice co-efficient}} \\
        \midrule
        Our model & 0.9001 & 0.7779 & 0.7152 & 0.8041 \\
        Arafat et al. \cite{b18} & 0.82 & 0.74 & 0.68 & 0.94 \\
        Gupta et al. \cite{b19} & 0.89 & 0.91 & {-} & 0.90 \\ 
        \bottomrule
    \end{tabular}
\end{table}
\section{Conclusion and future work}
The aim of this study was to systematically evaluate the impact of focal loss parameters and basic data augmentation strategies on U-Net–based brain tumor segmentation. Changes in focal loss parameters significantly impacted the model behavior, with better results obtained when parameters were tuned to give more weightage on minority class (tumor region) and hard-to-classify examples. Among the augmentation techniques, Horizontal flip was the most effective, followed by Rotation, while Scaling showed minimal improvement. These findings highlight the importance of careful loss function design and augmentation choice in developing robust segmentation pipelines.

Future work will extend this baseline study to include more advanced augmentation strategies such as elastic deformations, modality-specific transformations, etc. Artificially synthesized data using generative models like GANs can also be experimented with to evaluate its effect on model performance when used as an augmentation technique. Furthermore, classification of tumor type can also be integrated into the project pipeline such that when given an image, the model segments it and classifies the tumor into specific categories.
\vspace{\baselineskip}

\textit{\textbf{Data and code availability:}}
All code and experimental configurations are publicly available at github.com/Saumya4321/2d-brain-tumor-segmentation.

\vspace{12pt}

\end{document}